\documentclass[runningheads]{llncs}

\usepackage[T1]{fontenc}
\usepackage[utf8]{inputenc}
\usepackage{graphicx}
\usepackage{tabularx}
\usepackage{enumitem}
\usepackage{booktabs}
\usepackage{soul}
\usepackage{color}
\usepackage{tikz}
\usepackage{comment}

\newcolumntype{C}{>{\centering\arraybackslash}X}
\newcolumntype{R}{>{\raggedleft\arraybackslash}X}
\newcolumntype{L}{>{\raggedright\arraybackslash}X}

\begin{document}

\title{Lifelong Learning Natural Language Processing Approach for Multilingual Data Classification}
\titlerunning{Lifelong Learning Natural Language Processing \ldots}

\author{
	Jędrzej~Kozal\orcidID{0000-0001-7336-2561} \and
	Michał~Leś \and
	Paweł~Zyblewski\orcidID{0000-0002-4224-6709} \and
	Paweł~Ksieniewicz\orcidID{0000-0001-9578-8395} \and
	Michał~Woźniak\orcidID{0000-0003-0146-4205}
}

\authorrunning{J. Kozal et al.}

\institute{
	Wroclaw University of Science and Technology,\\
	Department of Systems and Computer Networks,\\
	Wroclaw, Poland\\
\email{jedrzej.kozal@pwr.edu.pl}}

\maketitle

\begin{abstract}

The abundance of information in digital media, which in today's world is the main source of knowledge about current events for the masses, makes it possible to spread disinformation on a larger scale than ever before. Consequently, there is a need to develop novel fake news detection approaches capable of adapting to changing factual contexts and generalizing previously or concurrently acquired knowledge. To deal with this problem, we propose a lifelong learning-inspired approach, which allows for fake news detection in multiple languages and the mutual transfer of knowledge acquired in each of them. Both classical feature extractors, such as \emph{Term frequency-inverse document frequency} or \emph{Latent Dirichlet Allocation}, and integrated deep \textsc{nlp} (\emph{Natural Language Processing}) \textsc{bert} (\emph{Bidirectional Encoder Representations from Transformers}) models paired with \textsc{mlp} (\emph{Multilayer Perceptron}) classifier, were employed. The results of experiments conducted on two datasets dedicated to the fake news classification task (in English and Spanish, respectively), supported by statistical analysis, confirmed that utilization of additional languages could improve performance for traditional methods. Also, in some cases supplementing the deep learning method with classical ones can positively impact obtained results.
The ability of models to generalize the knowledge acquired between the analyzed languages was also observed.

\end{abstract}

\section{Introduction}
It is almost a cliche to say that the modern economy and society strongly depend on information. Therefore we all expect that the information we are provided with will be reliable and credible, enabling us to make rational decisions. This is the ideal world, and the role of information and disinformation has been appreciated since ancient times and information manipulation has become a critical weapon used to gain political or material benefits~\cite{posetti2018short}. Nowadays, the problem with the use of so-called \emph{fake news} is strongly noticed because such a scale of manipulation has not been noted before. One can think of the news spread related to the COVID-19 pandemic or the Russian invasion of Ukraine.


This problem is particularly evident in digital media, hence almost all global Internet platforms, such as Facebook or Twitter, indicate in their terms of service the mechanisms for verifying information. Unfortunately, manual verification of the information veracity and relying on human experts as fact-checkers is relatively slow. \emph{MIT Media Lab} proves, that fake news travels farther and faster than the news that is absolutely legit\footnote{https://www.technologyreview.com/2018/03/08/144839/fake-news-spreads-faster-than-the-truth-and-its-all-our-fault/}. Hence, one of the challenges is to develop mechanisms that can detect fake news automatically and have the ability to improve their model continuously. 

Among the ML-based approaches to fake news detection, the literature distinguishes methods for \cite{choras2021advanced}:
\begin{itemize}
    \item Text analysis consists of analyzing the Natural Language Processing (\textsc{nlp}) data representation without the linguistic context \cite{saquete2020fighting}, as well as the psycholinguistic analysis \cite{zhang2016online} and processing the syntactic text structure \cite{kumar2019tree}. 
    \item Reputation analysis, which measures the reputation of an article and publisher based on sources such as content, reviews, domain, IP address or anonymity \cite{xu2019detecting}.
    \item Network analysis, which is related to the graph theory, and is performed to evaluate the truthfulness of the information \cite{zhou2019network}.
    \item Image-manipulation recognition, which are dedicated for the detection of image tampering \cite{bondi2017tampering}, copy-move forgeries \cite{bilal2020single}, and other image modifications such as contrast enhancement \cite{suryawanshi2019detection}.
\end{itemize}

This work deals with \textsc{nlp}, and introduces a training procedure for a fake news detection model that can classify news provided in different languages and perform knowledge transfer between them. 


\section{Related Works}

Fake news detection with \textsc{nlp} is based on the extraction of features directly from the content of the analyzed texts, without taking into account their social context, and is currently the basis for each of the subtasks included in the detection of fake news \cite{saquete2020fighting}.
Many of the classic \textsc{nlp} methods are based on \emph{bag-of-words} approach \cite{harris1954distributional}, which creates a vector for each document containing information about the number of times each word appears in it. An example of an extension of this approach may be $n$-grams, which can tokenize a sequence of words of a given length, rather than focusing only on individual words. The minimum range of $n$-grams is one (so-called unigrams), while its upper limit strongly depends on the length of the analyzed documents. Additionally, the weights of individual words can be normalized for a specific document using \emph{Term Frequency} (\textsc{tf}) \cite{luhn1957statistical}, or for the entire analyzed corpus, using \emph{Term Frequency-Inverse Document Frequency} (\textsc{tf-idf}) \cite{jones1972statistical}. Another noteworthy approach to extracting features from text is \emph{Latent Dirichlet Allocation} (\textsc{lda}), which is a generative probabilistic model for topic modeling \cite{lda}.

Such approaches, despite their simplicity, are successfully used in the literature in the task of recognizing \emph{fake news}. Hassan et al classified \emph{Tweets} as credible and not credible, using five different classification algorithms and word-based $n$-gram analysis including \textsc{tf} and \textsc{tf-idf} \cite{hassan2020credibility}. The authors have proven the effectiveness of this approach on the \textsc{pheme} benchmark dataset \cite{pheme}. Bharadwaj and Shao employed reccurent neural networks, random forest classifier and naive bayes classifier coupled with $n$-grams of various lengths in order to classify fake news dataset from \emph{Kaggle} \cite{bharadwaj2019fake}. Similar experiments, including e.g. $n$-grams, \textsc{tf-idf} and \emph{Gradient Boosting Classifier}, were conducted by Wynne and Wint \cite{wynne2019content}. Kaur et al. and Telang et al. evaluated multiple combinations of various classifier and extraction techniques in the fake news detection task \cite{kaur2020automating,telang2019anempirical}. Usage of the ensemble methods for disinformation detection was explored in 
\cite{ksieniewicz2019machine}.


Currently, deep neural networks \emph{Deep Neural Networks} (\textsc{dnn}) based on pre-training general language representations are very popular in fake news detecting. \cite{rodriguez2019fake}. 
A strong point of the currently used goings, which additionally fine-tune the token representations using supervised learning, is that a few parameters need to be learned from scratch only \cite{radford2018improving}. Recently, many works show the usefulness of knowledge transfer in the case of \emph{natural language inference} \cite{conneau2017supervised}.
\textsc{bert} (\emph{Bidirectional Encoder Representations from Transformers}) is particularly recognizable, which was designed for training with the use of unlabeled text \cite{devlin2018bert}. Then, \textsc{bert} is fine-tuned for a given problem with an additional output layer.

\section{Methods}

Each recognition system dedicated to language data is based on two basic steps. The first is vectorization, which allows for the transformation of a set of textual documents into a numerical data set interpretable by modeling procedure, and the second is an appropriate induction that allows for the construction of a model in a problem marked with bias. These procedures can be conducted separately, in a serial manner, as in the case of the classic \textsc{tf-idf} and \textsc{mlp} pair, where the basic extractor determines the vector representation and the neural network fits in the space designated by it, or jointly, in a parallel manner, as is the case in deep models, such as
\textsc{bert}.

Regardless of the integrated processing procedure typical of deep learning approaches, in which extraction and induction take place within the same structure, also in models such as \textsc{bert} we can delineate the constructive elements responsible for the extraction and -- at the end of the neural network structure -- fully connected layers, which in their logic are identical to typical, shallow \textsc{mlp} networks. This creates some potential for integrating classical and deep feature extraction methods for \textsc{nlp}. The deep model, getting trained on the basis of a given bias, trains weights responsible for both extraction of attributes and classification,
updating them so that they are better suited to the construction of the proper problem space. It is therefore possible to separate the extraction part of the deep learning model and -- using typical propagation -- use it to obtain attributes for the classification model in the same manner as classical extraction methods such as \textsc{tf-idf}.

Therefore, in the proposed method, the \textsc{mlp} model is used as a base classifier, the default in the structure for \emph{deep extractors}, and at the same time, good at optimizing the decision space for classical extraction methods. Moreover, for the analysis of the integration potential of heterogeneous extractors, three methods commonly used in the literature were selected:

\begin{itemize}
	\item \emph{Term Frequency-Inverse Document Frequency} -- legacy model based on normalization to document (\textsc{tf}) and to corpora (\textsc{idf}) as the most basic method currently used in \textsc{nlp} applications.
	\item \emph{Latent Dirichlet Allocation} -- a generative statistical model, similar to the latest unsupervised learning methods, but using approaches to identify significant thematic factors. 
	\item \emph{Bidirectional Encoder Representations from Transformers} -- the current state-of-the-art model, being a clear baseline in contemporary \textsc{nlp} experiments.	
\end{itemize}

The \textsc{bert} model, unlike \textsc{tf-idf} and \textsc{lda}, is not a language-independent \textsc{nlp} task solver. To enable its operation in a multilingual environment, base models were selected, trained on different corpora, with a similar structure size.
First was standard version pre-trained on English corpus (\textsc{bert}), second \textsc{beto}~\cite{CaneteCFP2020} - variant of \textsc{bert} architecture pre-trained on Spanish dataset (\textsc{bert}$_{spa}$), and third, the standard \textsc{bert} pre-trained on a multilingual corpus (\textsc{bert}$_{mult}$). 



\textsc{nlp} extractors, regardless of the procedure used, tend to generate many redundant attributes, significantly increasing the optimization space of the neural network. In the case of an integration architecture based on the concatenation of features obtained from a heterogeneous pool of extractors, this tendency will be further deepened.
Additionally, different extraction contexts may also allow for the diversification of models, where different processing algorithms allow for alternative insight on the data. For this proposal, the following solutions will be used:


\begin{itemize}
	\item \emph{Mutual Information} (\textsc{mi})
	\item \emph{Analysis of Variance} (\textsc{anova})
	\item \emph{Principal Components Analysis} (\textsc{pca})
\end{itemize}

\begin{figure}[!htb]
	\centering
	\resizebox{\textwidth}{!}{\begin{tikzpicture}[font=\sffamily]

\usetikzlibrary{shapes,snakes}
\tikzset{
    bases/.style = {draw, text width=3cm, inner sep=0pt, align=center, minimum height=.75cm},
    texts/.style = {text width=4cm, inner sep=0pt, align=center, minimum height=.75cm},
    texth/.style = {text width=10cm, inner sep=0pt, align=center, minimum height=.75cm},
    dia/.style = {diamond, draw, text width = 1cm, align = center,thick},
    ar/.style = {->, thick,shorten >=2.5pt},
    psi/.style = {bases, text width=1.5cm, fill=blue!15, minimum height=1cm, thick},
    ex/.style = {bases, text width=1.5cm, fill=red!15, minimum height=1cm, thick},
    exa/.style = {bases, text width=1.5cm, fill=green!15, minimum height=1cm, thick},
}

\draw[step=1,black,thin, dotted, white] (-1.5,2.5) grid (11.5,-11.5);

\node[texts, align=left] at (12.5,1.25) {\scriptsize\emph{corpora A}};
\node[texts, align=left] at (12.5,1) {\scriptsize\emph{corpora B}};
\node[texts, align=left] at (12.5,.75) {\scriptsize\emph{corpora C}};	

\draw[ar] (9.75,-1) -- (9.75,.65) -- (10.5,1.25);
\draw[ar] (9.85,-1) -- (9.85,.5) -- (10.5,1);
\draw[ar] (9.95,-1) -- (9.95,.35) -- (10.5,.75);

\node[texth, align=center, rotate=90, text width=5cm] at (-1.125,-2.5) {\bfseries\emph{Simple integration schema}};

\foreach \x in {1,3,...,9}{
	\foreach \d in {.1,.2,.3}{
		\node[ex] at (\x-\d+.3,-1+\d-.3) {Extractor};
		\node[psi] at (\x-\d+.3,-2.5+\d-.3) {$\Psi$};
	}
} 

\foreach \d in {.1,.2,.3}{
	\node[psi, text width=9.5cm] at(5-\d+.3,-4+\d-.3) {$Ensemble$};
}

\node[texts, align=center] at (1,.5) {\bfseries TF-IDF};
\node[texts, align=center] at (3,.5) {\bfseries LDA};
\node[texts, align=center] at (5,.5) {\bfseries MULT};
\node[texts, align=center] at (7,.5) {\bfseries ENG};
\node[texts, align=center] at (9,.5) {\bfseries SPA};
\node[texts, align=center] at (7,1.5) {\bfseries BERT};

\draw[black, ultra thick] (7,1.25) -- (7,.75);
\draw[black, ultra thick] (5,.75)-- (5,1) -- (9,1) -- (9,.75);

\node[texth, align=center, rotate=90] at (-1.125,-8.5) {\bfseries\emph{Extractor reduction schema}};

\foreach \x in {1,3,...,9}{
	\foreach \d in {.1,.2,.3}{
		\node[ex] at (\x-\d+.3,-6+\d-.3) {Extractor};
		\draw[ar] (\x,-1.5) -- (\x,-2);
		\draw[ar] (\x,-3) -- (\x,-3.5);
	}
} 
\foreach \x in {0,1,...,4}{
	\draw[ar] (.75+2*\x-.75+1,-6.5) -- (2+1.5*\x,-7);
}

\foreach \d in {.1,.2,.3}{
	\node[exa, text width=7.5cm] at(5-\d+.3,-7.5+\d-.3) {Concatenated feature space};
	\foreach \x in {0,1.5,...,4.5}{
		\draw[dashed, black!50] (2.75+\x-\d+.3,-7+\d-.3) -- (2.75+\x-\d+.3,-8+\d-.3);
	}
	\node[ex, text width=1.5cm] at(5-\d+.3,-9+\d-.3) {Reductor};
	\node[psi, text width=1.5cm] at(5-\d+.3,-10.5+\d-.3) {$\Psi$};
}
\draw[ar] (5,-8) -- (5,-8.5);
\draw[ar] (5,-9.5) -- (5,-10);

\node[texth, align=center] at (5,2) {\emph{Feature extraction approaches}};

\draw[thick] (-1.5,-5) -- (11.5,-5);
\draw[thick] (-1.5,0) -- (11.5,0);

\end{tikzpicture}}
	\caption{Ensemble diversification policies}\label{fig:schema}
\end{figure}
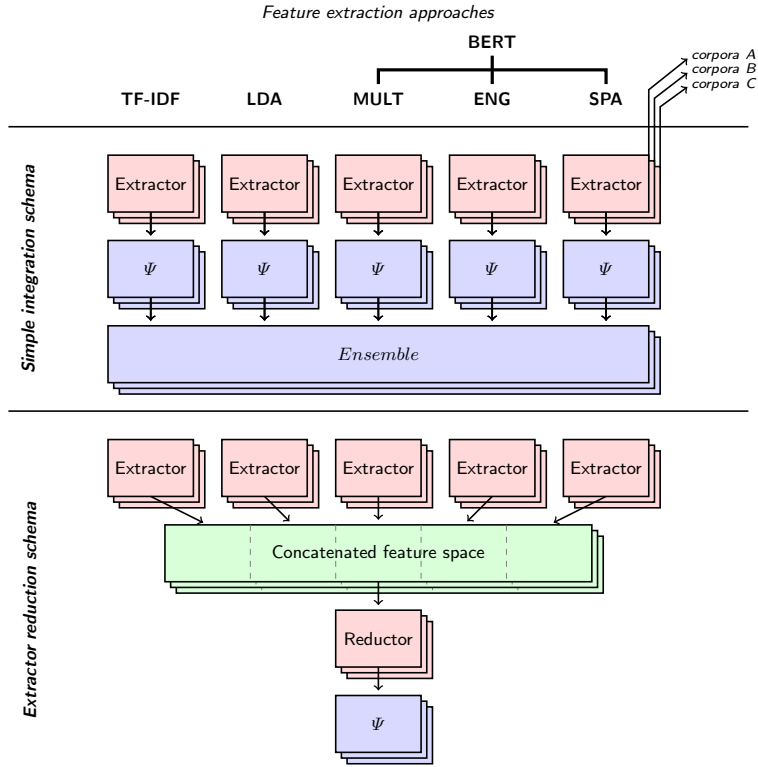

With the use of all the tools introduced above, it is possible to propose two alternative architectures for integrating the extractors. They are presented in Figure~\ref{fig:schema}, compiling the processing schemes of \textsc{sis} and \textsc{ers} policies.

As part of the \emph{Simple Integration Schema} policy (\textsc{sis}), each of the extractors -- independently -- builds an extraction model which, after processing the available language corpora (\emph{A, B and C corpora}) for each  of them, allows for the transformation from a text into a vector representation by pre-limited, constant dimensionality. For each transformation, a separate classifier is built, which finally goes to the pools of models dedicated to different corpora, integrated internally and externally through the accumulation of support. Internal integration is carried out classically, and external integration depends on the source languages of the corpora. In the case of a linguistically homogeneous pool, the accumulation of support is not modified. Otherwise, (\emph{a}) the number of languages is a multiplier of the number of problem classes, (\emph{b}) monolingual models receive zero foreign language support, and (\emph{c}) multilingual models replicate class supports divided by the language multiplier to preserve the complementarity condition of the support vector.

The \emph{Extractor Reduction Schema}  
(\textsc{ers}) strategy performs the extraction in the same way for \textsc{sis}. However, models are not built for each of the extractors, and the object representations obtained within the corpora are concatenated into a wide, common space of the problem. 
The space dimensionality is reduced to a predetermined size, i.e.
upper limit of the number of attributes per extractor. A separate classification model is trained for each corpora, constituting a narrow pool that is integrated in the same way as external \textsc{sis} ensemble policy.

The aim of the research is to obtain experimentally verified information about the appropriate integration strategy of classifier ensembles for the needs of multilingual natural language processing systems. Both the potentials of various extraction methods -- in each of the selected corpora -- and the capabilities of the various dimension reducers in the \textsc{ers} policy will be verified.

\section{Experiment}



The experiments aim to analyze how knowledge gained from different languages can positively affect the overall classification performance. During the experimental evaluation, we intend to answer the following research questions:

\begin{enumerate}
    \item How does fake news classification performance depend on different extraction methods?
    \item How do ensemble models diversified by various extractors, training sets and training features improve fake-news classification?
    \item Can we utilize models trained with different language corpora to improve other language classification?
    \item What methods for feature integration can be useful in a multilingual environment?

\end{enumerate}

\subsection{Setup}

\noindent \textbf{Datasets.} Two datasets were utilized in this study, namely: \emph{The Spanish Fake News Corpus} \footnote{https://github.com/jpposadas/FakeNewsCorpusSpanish} \cite{esp_fake} and \emph{Kaggle Fake News: Build a system to identify unreliable news articles}\footnote{https://www.kaggle.com/c/fake-news/data}. These datasets were treated as a representation of fake-news distribution in Spanish and English, respectively. The third variant of a dataset was a mixed dataset that combined both Spanish and English datasets. Detailed information about the number of samples for each dataset is given in Table~\ref{tab:datasets}. As we can see both Spanish and English datasets are internally balanced. However, please note a significant, 1:31 imbalance in the presence of the source datasets in \emph{mixed} article collection. For this reason in further parts we report balanced accuracy as the evaluation metric, which in case of \emph{mixed} dataset actually shows the quality of the four-class problem -- sensitizing itself to errors performed on minority language.

\begin{table}[!htb]
	\centering
	\caption{
	Description of used datasets. Fake-news learning examples are considered as positives.}
	\label{tab:datasets}
	\begin{tabularx}{0.8\textwidth}{lRRRC}
	\toprule
	\bfseries\textsc{dataset} 
	& \bfseries\textsc{\#samples} 
	& \bfseries\textsc{\#negative}
	& \bfseries\textsc{\#positive}
	& \emph{ir}  \\
	\midrule
	\emph{esp fake} & 676 & 338 & 338 & 1:1 \\
	\emph{kaggle} & 20 800 & 10 387 & 10 413 & 1:1 \\
	\midrule
	\emph{mixed} & 21 476 & 10 725 & 10 751 & 1:1\\\bottomrule
	\emph{ir} & 1:31 & 1:31 & 1:31 &\\
\end{tabularx}
\end{table}

\noindent \textbf{Data preprocessing.} Each word was changed to lowercase before computing \textsc{tf-idf} features. For this extraction method, features were computed based on unigrams. We remove words that occur in less than 20 documents or more than half of all documents during \textsc{lda} preprocessing. \textsc{lda} features were computed with chunk size 2000, 20 passes, and 400 iterations. Both \textsc{tf-idf} and \textsc{lda} extracted 100 features from each document. 

\noindent \textbf{Classification models.} MLP used for training on extracted features had two hidden layers with 500 neurons each and was trained by Adam optimizer \cite{adam} with a learning rate 0.001 for 200 epochs with Early Stopping \cite{early_stopping}. \textsc{bert}-based models were trained for 5 epochs each with AdamW \cite{DBLP:journals/corr/abs-1711-05101} optimizer and learning rate 3e-5.

 \noindent \textbf{Experimental protocol.} To obtain more reliable results 5x2 cross validation was employed. However, there is a significant difference in the number of English and Spanish datasets samples. For this reason, stratification was introduced for Spanish and English labels in the case of the mixed dataset. This modification allows avoiding over-representation of learning examples from one language in some folds. 

\noindent \textbf{Result analysis.} To analyse the classification performance of the chosen algorithms  we employed a 5x2cv combined F test for statistical analysis with p-value of 0.05.

\noindent \textbf{Implementation and reproducibility.} Github repository with code for all experiments is available online 
\footnote{https://github.com/w4k2/nn-nlp}. 

\subsection{Results}

\subsubsection{E1 Classification based on single feature extraction method}

Results of classification accuracy obtained for a single feature extraction methods are presented in Tables~\ref{tab:1Mtext} and \ref{tab:1Mtitle}. We use these values as a baseline for further experiments. We state a significant difference between all deep models and \textsc{tf-idf} and \textsc{lda} for text and title attributes. Additionally, for text attribute \textsc{bert} was better than \textsc{bert}$_{mult}$ for the \emph{kaggle} dataset, \textsc{lda} than \textsc{tf-idf} for the \emph{kaggle} and \emph{mixed} datasets. Solution based on \textsc{bert}$_{spa}$ was better than \textsc{bert}$_{mult}$ for \emph{esp fake} dataset and better than any other model for the \emph{mixed} dataset. 

\begin{table}
	\caption{Accuracy for each extractor trained and evaluated on one dataset for text attribute.}
	\label{tab:1Mtext}
	\begin{tabularx}{0.99\textwidth}{l|C|C|C|C|C}
		\toprule
		\bfseries\textsc{dataset} & \multicolumn{5}{c}{\bfseries\textsc{extraction method}}\\
		& & & \multicolumn{3}{c}{\bfseries\textsc{bert}}\\
		& \bfseries\textsc{tf-idf} & \bfseries\textsc{lda} & $mult$& $eng$ & $spa$\\
		& \bfseries \oldstylenums{1}& \bfseries \oldstylenums{2}& \bfseries \oldstylenums{3}& \bfseries \oldstylenums{4}& \bfseries \oldstylenums{5}\\\midrule
		
		\emph{kaggle} &  .866 &  .911 &  .981 &  .985 &  ---  \\
        & \scriptsize ---  & \scriptsize 1   & \scriptsize 1, 2   & \scriptsize 1, 2, 3   & \scriptsize ---   \\
        \emph{esp fake} &  .751 &  .618 &  .786 &  --- &  .853  \\
        & \scriptsize 2   & \scriptsize ---  & \scriptsize 2   & \scriptsize ---  & \scriptsize 1, 2, 3    \\
        \emph{mixed} &  .729 &  .768 &  .972 &  .974 &  .982  \\
        & \scriptsize ---  & \scriptsize 1   & \scriptsize 1, 2   & \scriptsize 1, 2   & \scriptsize 1, 2, 3, 4    \\
		\bottomrule
	\end{tabularx}
\end{table}

\begin{table}[!htb]
	\caption{Accuracy for each extractor trained and evaluated on one dataset for title attribute.}
	\label{tab:1Mtitle}
	\begin{tabularx}{0.99\textwidth}{l|C|C|C|C|C}
	\toprule
		\bfseries\textsc{dataset} & \multicolumn{5}{c}{\bfseries\textsc{extraction method}}\\
		& & & \multicolumn{3}{c}{\bfseries\textsc{bert}}\\
		& \bfseries\textsc{tf-idf} & \bfseries\textsc{lda} & \bfseries\textsc{mult}& \bfseries\textsc{eng} & \bfseries\textsc{spa}\\
		& \bfseries \oldstylenums{1}& \bfseries \oldstylenums{2}& \bfseries \oldstylenums{3}& \bfseries \oldstylenums{4}& \bfseries \oldstylenums{5}\\\midrule
		
        \emph{kaggle} &  .904 &  .829 &  .958 &  .963 &  ---  \\
        & \scriptsize 2  & \scriptsize --- & \scriptsize 1, 2  & \scriptsize 1, 2  & \scriptsize ---   \\
        \emph{esp fake} &  .540 &  .553 &  .675 &  --- &  .852  \\
        & \scriptsize --- & \scriptsize --- & \scriptsize 1, 2  & \scriptsize --- & \scriptsize 1, 2, 3    \\
        \emph{mixed} &  .646 &  .631 &  .942 &  .948 &  .982  \\
        & \scriptsize --- & \scriptsize --- & \scriptsize 1, 2  & \scriptsize 1, 2, 3  & \scriptsize 1, 2, 3, 4    \\
		\bottomrule
	\end{tabularx}
\end{table}

\subsubsection{E2 Classification based on multiple feature extraction methods}

In this part of experiments we create ensemble according to \emph{Simple Integration Schema} and \emph{Extractor Reduction Schema} introduced earlier. We use multiple base models trained with the same language.
Results are presented in tables \ref{tab:4Mtext} and \ref{tab:4Mtitle}. \textsc{pca} was significantly worse than \textsc{sa} for esp fake and mixed and all other methods for the mixed dataset with text attribute. For title attribute, \textsc{pca} was better than \textsc{sa} with the kaggle dataset, \textsc{sa} was better than all other methods for esp fake, \textsc{pca} was worse than other methods, and \textsc{minfo} and \textsc{anova} were better than \textsc{sa} for the mixed dataset.

\begin{table}[!htb]
	\caption{Accuracy for different ensemble construction methods with text attribute.}
	\label{tab:4Mtext}
	\begin{tabularx}{0.99\textwidth}{l|C|C|C|C}
		\toprule
		\bfseries\textsc{dataset} & \multicolumn{4}{c}{\bfseries\textsc{integration method}}\\
		& & \multicolumn{3}{c}{\bfseries\textsc{dimensionality reduction}}\\
		& \bfseries\textsc{sa} & \bfseries\textsc{minfo} & \bfseries\textsc{anova}& \bfseries\textsc{pca} \\
		& \bfseries \oldstylenums{1}& \bfseries \oldstylenums{2}& \bfseries \oldstylenums{3}& \bfseries \oldstylenums{4} \\\midrule
		
        \emph{kaggle} &  .988 &  .987 &  .987 &  .983  \\
        & \scriptsize --- & \scriptsize --- & \scriptsize --- & \scriptsize ---   \\
        \emph{esp fake} &  .851 &  .833 &  .831 &  .788  \\
        & \scriptsize 4  & \scriptsize --- & \scriptsize --- & \scriptsize ---   \\
        \emph{mixed} &  .913 &  .894 &  .903 &  .785  \\
        & \scriptsize 4  & \scriptsize 4  & \scriptsize 4  & \scriptsize ---   \\
		\bottomrule
	\end{tabularx}
\end{table}

\begin{table}[!htb]
	\caption{Accuracy for different ensemble construction methods with title attribute.}
	\label{tab:4Mtitle}
		\begin{tabularx}{0.99\textwidth}{l|C|C|C|C}
		\toprule
		\bfseries\textsc{dataset} & \multicolumn{4}{c}{\bfseries\textsc{integration method}}\\
		& & \multicolumn{3}{c}{\bfseries\textsc{dimensionality reduction}}\\
		& \bfseries\textsc{sa} & \bfseries\textsc{minfo} & \bfseries\textsc{anova}& \bfseries\textsc{pca} \\
		& \bfseries \oldstylenums{1}& \bfseries \oldstylenums{2}& \bfseries \oldstylenums{3}& \bfseries \oldstylenums{4} \\\midrule
		
        \emph{kaggle} &  .955 &  .963 &  .963 &  .964  \\
        & \scriptsize --- & \scriptsize 1, 3  & \scriptsize 1  & \scriptsize 1    \\
        \emph{esp fake} &  .730 &  .737 &  .732 &  .674  \\
        & \scriptsize 4  & \scriptsize --- & \scriptsize --- & \scriptsize ---   \\
        \emph{mixed} &  .885 &  .905 &  .879 &  .760  \\
        & \scriptsize 4  & \scriptsize 4  & \scriptsize 4  & \scriptsize ---   \\
		\bottomrule
	\end{tabularx}
\end{table}

\subsubsection{E3 Classification with different languages}

We utilize a single extraction method trained on different datasets to precompute features for one dataset. Extracted features are concatenated. Then, these features are used to train the single classifier. 
The impact of different language features on fake news classification performance is presented in table \ref{tab:3Mtext} and \ref{tab:3Mtitle}. To keep the presentation of the results consistent, the statistical analysis was carried out by comparing different models while keeping the dataset and ensemble construction method the same. No significant difference was found for text or title attributes.

\begin{table}[!htb]
	\caption{Accuracy for ensembles obtained from evaluating one extractor on all 3 datasets with text attribute.}
	\label{tab:3Mtext}
	\begin{tabularx}{\textwidth}{p{8em}|C|C|C|C|C}
	\toprule
	
	\bfseries\textsc{dataset} &
	\multicolumn{5}{c}{\bfseries \textsc{vectorization approach}}
	\\  
	
	& & & \multicolumn{2}{c|}{\bfseries\textsc{bert}} & 
	
	\\
	&
	\bfseries\textsc{tf-idf} &
	\bfseries\textsc{lda}  &
	\bfseries\textsc{mult.}  & 
	\bfseries\textsc{eng.}  &  
	\bfseries\textsc{beto}   
	\\
	\midrule	
	\multicolumn{6}{c}{\bfseries\textsc{SA}} \\
	\midrule
	\emph{kaggle} &  .933 &  .951 &  .934 &  .950 &  ---   \\
    & \scriptsize --- & \scriptsize --- & \scriptsize --- & \scriptsize ---  & \scriptsize ---   \\
    \emph{esp fake} &  .741 &  .764 &  .757 &  ---  &  .761  \\
    & \scriptsize --- & \scriptsize --- & \scriptsize ---  & \scriptsize --- & \scriptsize ---   \\
    \emph{mixed} &  .775 &  .772 &  .782 &  .740 &  .780  \\
    & \scriptsize --- & \scriptsize --- & \scriptsize --- & \scriptsize --- & \scriptsize ---   \\
	\midrule
	\multicolumn{6}{c}{\bfseries\textsc{MINFO}} \\
	\midrule
	\emph{kaggle} &  .922 &  .943 &  .921 &  .941 &  ---   \\
    & \scriptsize --- & \scriptsize --- & \scriptsize --- & \scriptsize ---  & \scriptsize ---   \\
    \emph{esp fake} &  .759 &  .772 &  .781 &  ---  &  .746  \\
    & \scriptsize --- & \scriptsize --- & \scriptsize ---  & \scriptsize --- & \scriptsize ---   \\
    \emph{mixed} &  .809 &  .802 &  .805 &  .801 &  .818  \\
    & \scriptsize --- & \scriptsize --- & \scriptsize --- & \scriptsize --- & \scriptsize ---   \\
	\midrule
	\multicolumn{6}{c}{\bfseries\textsc{ANOVA}} \\
	\midrule
	\emph{kaggle} &  .924 &  .945 &  .925 &  .944 &  ---   \\
    & \scriptsize --- & \scriptsize --- & \scriptsize --- & \scriptsize ---  & \scriptsize ---   \\
    \emph{esp fake} &  .771 &  .773 &  .783 &  ---  &  .748  \\
    & \scriptsize --- & \scriptsize --- & \scriptsize ---  & \scriptsize --- & \scriptsize ---   \\
    \emph{mixed} &  .801 &  .793 &  .795 &  .773 &  .791  \\
    & \scriptsize --- & \scriptsize --- & \scriptsize --- & \scriptsize --- & \scriptsize ---   \\
	\midrule
	\multicolumn{6}{c}{\bfseries\textsc{PCA}} \\
	\midrule
	\emph{kaggle} &  .921 &  .942 &  .922 &  .944 &  ---   \\
    & \scriptsize --- & \scriptsize --- & \scriptsize --- & \scriptsize ---  & \scriptsize ---   \\
    \emph{esp fake} &  .757 &  .775 &  .780 &  ---  &  .749  \\
    & \scriptsize --- & \scriptsize --- & \scriptsize ---  & \scriptsize --- & \scriptsize ---   \\
    \emph{mixed} &  .790 &  .815 &  .782 &  .788 &  .813  \\
    & \scriptsize --- & \scriptsize --- & \scriptsize --- & \scriptsize --- & \scriptsize ---   \\
	\bottomrule
\end{tabularx}

\end{table}

\begin{table}[!htb]
	\caption{Accuracy for ensembles obtained from evaluating one extractor with all 3 datasets with title attribute.}
	\label{tab:3Mtitle}
	\begin{tabularx}{\textwidth}{p{8em}|C|C|C|C|C}
\toprule

\bfseries\textsc{dataset} &
\multicolumn{5}{c}{\bfseries \textsc{vectorization approach}}
\\  

& & & \multicolumn{2}{c|}{\bfseries\textsc{bert}} & 

\\
&
\bfseries\textsc{tf-idf} &
\bfseries\textsc{lda}  &
\bfseries\textsc{mult.}  & 
\bfseries\textsc{eng.}  &  
\bfseries\textsc{beto}   
\\
\midrule
\multicolumn{6}{c}{\bfseries \textsc{SA}} \\\midrule
\emph{kaggle} &  .916 &  .931 &  .913 &  .928 &  ---   \\
& \scriptsize --- & \scriptsize --- & \scriptsize --- & \scriptsize ---  & \scriptsize ---   \\
\emph{esp fake} &  .638 &  .668 &  .655 &  ---  &  .622  \\
& \scriptsize --- & \scriptsize --- & \scriptsize ---  & \scriptsize --- & \scriptsize ---   \\
\emph{mixed} &  .676 &  .688 &  .678 &  .659 &  .669  \\
& \scriptsize --- & \scriptsize --- & \scriptsize --- & \scriptsize --- & \scriptsize ---   \\
\midrule
\multicolumn{6}{c}{\bfseries\textsc{MINFO}} \\
\midrule
\emph{kaggle} &  .903 &  .921 &  .901 &  .921 &  ---   \\
& \scriptsize --- & \scriptsize --- & \scriptsize --- & \scriptsize ---  & \scriptsize ---   \\
\emph{esp fake} &  .628 &  .680 &  .681 &  ---  &  .635  \\
& \scriptsize --- & \scriptsize --- & \scriptsize ---  & \scriptsize --- & \scriptsize ---   \\
\emph{mixed} &  .740 &  .743 &  .747 &  .742 &  .739  \\
& \scriptsize --- & \scriptsize --- & \scriptsize --- & \scriptsize --- & \scriptsize ---   \\
\midrule
\multicolumn{6}{c}{\bfseries\textsc{ANOVA}} \\
\midrule
\emph{kaggle} &  .914 &  .926 &  .901 &  .923 &  ---   \\
& \scriptsize --- & \scriptsize --- & \scriptsize --- & \scriptsize ---  & \scriptsize ---   \\
\emph{esp fake} &  .633 &  .682 &  .685 &  ---  &  .627  \\
& \scriptsize --- & \scriptsize --- & \scriptsize ---  & \scriptsize --- & \scriptsize ---   \\
\emph{mixed} &  .751 &  .756 &  .750 &  .749 &  .743  \\
& \scriptsize --- & \scriptsize --- & \scriptsize --- & \scriptsize --- & \scriptsize ---   \\
\midrule
\multicolumn{6}{c}{\bfseries\textsc{PCA}} \\
\midrule
\emph{kaggle} &  .913 &  .929 &  .911 &  .927 &  ---   \\
& \scriptsize --- & \scriptsize --- & \scriptsize --- & \scriptsize ---  & \scriptsize ---   \\
\emph{esp fake} &  .618 &  .685 &  .679 &  ---  &  .631  \\
& \scriptsize --- & \scriptsize --- & \scriptsize ---  & \scriptsize --- & \scriptsize ---   \\
\emph{mixed} &  .764 &  .756 &  .769 &  .727 &  .767  \\
& \scriptsize --- & \scriptsize --- & \scriptsize --- & \scriptsize --- & \scriptsize ---   \\
\bottomrule
\end{tabularx}

\end{table}

\subsubsection{E4 Classification with different extraction methods and languages}

Classification performance of ensemble using a combination of all languages and extraction methods is presented in tables \ref{tab:12Mtext} and \ref{tab:12Mtitle}. For text attribute, we state meaningful differences between \textsc{pca} and the rest of the methods and between \textsc{minfo} and \textsc{sa} for the kaggle dataset and the difference between ANVOA and \textsc{pca} for the mixed dataset. For title attribute, we found that \textsc{pca} is better than \textsc{minfo} for kaggle, \textsc{sa} is better than \textsc{pca} for esp fake. For mixed dataset, \textsc{anova} is better than all other methods, \textsc{sa} is better than \textsc{pca}, and \textsc{minfo} is better than \textsc{sa} and \textsc{pca}.

\begin{table}[!htb]
	\caption{Results for utilizing in ensemble both different extraction methods and different languages with text attribute.}
	\label{tab:12Mtext}
	\begin{tabularx}{0.99\textwidth}{l|C|C|C|C}
    \toprule
    \bfseries\textsc{dataset} & \multicolumn{4}{c}{\bfseries\textsc{integration method}}\\
    & & \multicolumn{3}{c}{\bfseries\textsc{dimensionality reduction}}\\
    & \bfseries\textsc{sa} & \bfseries\textsc{minfo} & \bfseries\textsc{anova}& \bfseries\textsc{pca} \\
    & \bfseries \oldstylenums{1}& \bfseries \oldstylenums{2}& \bfseries \oldstylenums{3}& \bfseries \oldstylenums{4} \\\midrule

    \emph{kaggle} &  .986 &  .988 &  .988 &  .990  \\
    & \scriptsize --- & \scriptsize --- & \scriptsize --- & \scriptsize 1    \\
    \emph{esp fake} &  .845 &  .830 &  .831 &  .804  \\
    & \scriptsize --- & \scriptsize --- & \scriptsize --- & \scriptsize ---   \\
    \emph{mixed} &  .884 &  .875 &  .838 &  .869  \\
    & \scriptsize 3  & \scriptsize --- & \scriptsize --- & \scriptsize ---   \\
\bottomrule
\end{tabularx}
\end{table}

\begin{table}[!htb]
	\caption{Results for utilizing in ensemble both different extraction methods and different languages with title attribute.}
	\label{tab:12Mtitle}
	\begin{tabularx}{0.99\textwidth}{l|C|C|C|C}
	\toprule
	\bfseries\textsc{dataset} & \multicolumn{4}{c}{\bfseries\textsc{integration method}}\\
	& & \multicolumn{3}{c}{\bfseries\textsc{dimensionality reduction}}\\
	& \bfseries\textsc{sa} & \bfseries\textsc{minfo} & \bfseries\textsc{anova}& \bfseries\textsc{pca} \\
	& \bfseries \oldstylenums{1}& \bfseries \oldstylenums{2}& \bfseries \oldstylenums{3}& \bfseries \oldstylenums{4} \\\midrule
	
    \emph{kaggle} &  .965 &  .963 &  .964 &  .963  \\
    & \scriptsize --- & \scriptsize --- & \scriptsize --- & \scriptsize ---   \\
    \emph{esp fake} &  .771 &  .750 &  .753 &  .673  \\
    & \scriptsize 4  & \scriptsize --- & \scriptsize --- & \scriptsize ---   \\
    \emph{mixed} &  .762 &  .908 &  .909 &  .806  \\
    & \scriptsize --- & \scriptsize 1, 4  & \scriptsize 1, 4  & \scriptsize ---   \\
	\bottomrule
	\end{tabularx}
\end{table}

\subsubsection{Visualization of learned features}

\begin{figure}[!htb]
    \centering
    \textsc{lda}
    \includegraphics[width=\textwidth,clip=true,trim=0 280 0 0]{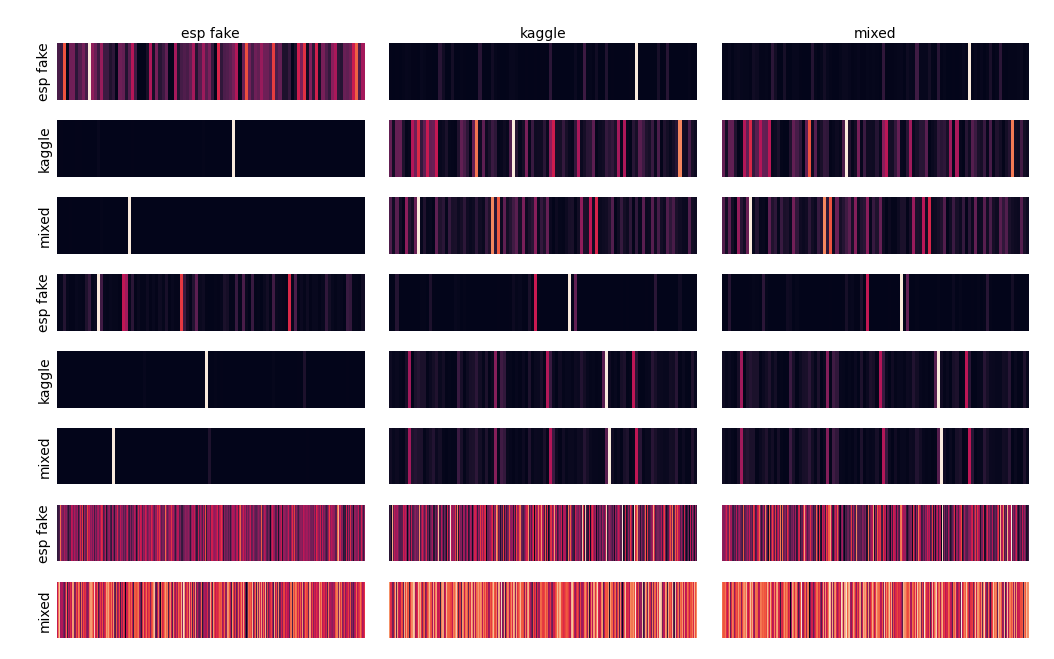}
    \textsc{tf-idf}
    \includegraphics[width=\textwidth,clip=true,trim=0 120 0 190]{images/heatmap3.png}
    \textsc{beto}
    \includegraphics[width=\textwidth,clip=true,trim=0 0 0 350]{images/heatmap3.png}
    \caption{Visualization of average features learned by \textsc{lda} and \textsc{tf-idf} and \textsc{beto}. Dataset labels in y-axis correspond to training datasets and labels in x-axis correspond to evaluation datasets. Brighter colors correspond to higher values.}
    \label{fig:avrg_features}
\end{figure}

To provide more insight into learned representations, we visualize average feature vectors. We consider features for the dataset that the extractor was trained on and features generated by applying the trained extractor to two other datasets. Next, we plot the average feature vector as a heatmap to show the most active features for each training-evaluation dataset pair. Results are presented in Fig. \ref{fig:avrg_features}. First, we focus on plots for \textsc{tf-idf} and \textsc{lda}. Most active features are obtained for the same dataset that the extractor was trained on. When applying the extractor to the datasets with a different language, some features have significantly higher values.
This indicates that applying the extraction model to a different language can provide an additional source of information that can be beneficial for classification quality. Furthermore, it can explain why exploiting extractors trained on different languages improved performance for \textsc{tf-idf} and \textsc{lda} in the third experiment. 

These findings are strictly connected to the inner workings of extraction algorithms. When extracting features with \textsc{tf-idf} first step is to create vocabulary from the corpus. Next, feature vectors are computed based on the frequency of each word. When processing text with words that mostly fall outside of vocabulary, \textsc{tf-idf} will return a sparse vector. Similar reasoning can be applied to \textsc{lda}. 
Transformers use a different procedure for text preprocessing. \textsc{bert} tokenizer can divide a single word into separate tokens if it falls outside of vocabulary. Then these tokens are used to generate word embeddings that are passed through the transformer network. As a result, when dealing with different languages model will not produce sparse output but some vector that may or may not contain useful information. This is shown in the two lowest rows of Fig \ref{fig:avrg_features}. There is no significant difference between \textsc{beto} average features, regardless of whether they are computed with the same language model was trained on or not. This in turn, can explain why we observe a decrease in accuracy for the deep models in Experiment 3.

\subsection{Lessons learned}

When analyzing accuracy obtained for single extractors (Tab. \ref{tab:1Mtext} and \ref{tab:1Mtitle}) one can notice that deep learning-based methods obtain best results. This is expected, as deep models currently dominate \textsc{nlp} field \cite{DBLP:journals/corr/abs-1810-04805,gpt2,DBLP:journals/corr/abs-2005-14165}. 
One can also notice that models dedicated for a specific language obtained better results than MultiBERT. This can be explained by a lower number of learning examples in datasets utilized in our experiments. MultiBERT is the largest model. Therefore, it can be prone to overfitting with a smaller amount of data. 
Also, it is worth noting that the difference in performance between \textsc{beto} and MultiBERT is larger than between the English version of \textsc{bert} and MultiBERT. This is probably caused by an improved pretraining procedure from RoBERTa \cite{DBLP:journals/corr/abs-1907-11692}, that was used in \textsc{beto}. Better pretraining can lead to improved performance in downstream tasks \cite{DBLP:journals/corr/abs-1810-04805,gpt2}.
Although MultiBERT was initially trained with a multi-language corpus and has the largest number of parameters, it is not the best model for the mixed dataset. Unfortunately, we found no good explanation for this phenomenon.
Results obtained for the Spanish dataset are worse compared to kaggle and mixed. This can be easily explained by a larger number of samples in the kaggle dataset compared to esp fake (please refer to Tab. \ref{tab:datasets}). These findings provide an answer to research question 1.
Deep models obtain better or similar results when comparing text and title attributes. Unfortunately, the same cannot be stated about \textsc{tf-idf} and \textsc{lda}, where performance for the title attribute is worse in most cases.

Regardless of the ensemble construction methodology obtained accuracy is close for all methods (Tab. \ref{tab:4Mtext} and \ref{tab:4Mtitle}). 
Comparing values of metrics to the first experiment, we can state that employing an ensemble can benefit accuracy only for the English dataset. This is probably caused by the close performance of all models for the kaggle dataset. In the case of the Spanish dataset, there is a clear gap between \textsc{beto} and the rest of the models, and in the case of the mixed dataset, there is a gap between deep learning models and two other. When combining strong classifiers with weaker ones in one ensemble, we can obtain worse results than for the strongest classifier. In this case, the utilization of additional models with worse performance is more harmful than helpful. These results answer research question 2.

In the third experiment feature extractor trained on a single dataset was used to extract features from all datasets and create a new ensemble from 3 different sets of features (Tab. \ref{tab:3Mtext} and \ref{tab:3Mtitle}). This approach is beneficial for \textsc{tf-idf} and \textsc{lda} but detrimental for deep models. The statistical analysis results show no significant differences between \textsc{tf-idf}, \textsc{lda}, and deep models. This finding is quite interesting due to the clear dominance of deep models in the first experiment. It is important to note that deep models achieved worse results than in the first experiment; however, at the same time, there was a significant gain in performance for \textsc{tf-idf} and \textsc{lda}. We provide a possible explanation when discussing feature visualization from Fig \ref{fig:avrg_features}. This answers research question 3.

Utilizing both different models and features extracted for different languages (Tab. \ref{tab:12Mtext} and \ref{tab:12Mtitle}) does not provide an improvement over baselines, nor constructing ensembles with the utilization of different models. These findings, combined with conclusions from Experiment 2, answer research question 4.

\section{Conclusions}
This study aimed to apply \textsc{nlp} methods to the problem of detecting misinformation in messages produced in different languages and verify whether it is possible to transfer knowledge between models trained for different languages. 


Based on the results of the experimental studies, it was not found that different models or attributes extracted for different languages led to a noticeable, i.e., statistically significant, improvement over the baseline results. Also, no significant improvement was confirmed using methods based on the classifier ensemble concept. The limited scope of datasets contents can explain this. Articles in the Spanish dataset are written about local affairs. Therefore intersections of topics between two datasets can be small. Moreover, this indicates that preparing a fake-news detection model with a single language and utilizing learned knowledge for other languages is a difficult problem and should be further explored in research. Having this in mind, future work should focus on preparing articles and datasets for multiple languages. By utilizing the same type of article topics in multiple languages, one can obtain better results. Our work showed that some classes of models could benefit from utilizing multiple languages, so it is reasonable to expect that aligning of topics should further improve results.

In the calculation of the achieved results, as well as the literature analysis, it seems that the direction of further work may be related to the transfer of knowledge, but rather within a single language, while it seems attractive to transfer knowledge between tasks of fake news detection, but different subject areas.

\bibliographystyle{splncs04}
\bibliography{bibliography}

\end{document}